\documentclass[10pt,journal,compsoc]{IEEEtran}

\ifCLASSOPTIONcompsoc
  \usepackage[nocompress]{cite}
\else
  \usepackage{cite}
\fi

\usepackage{gensymb}
\usepackage{mathtools}
\usepackage{rotating}

\usepackage{booktabs}
\usepackage{subcaption}
\usepackage{algorithm}
\usepackage{algorithmic}
\usepackage{enumitem}
\usepackage{multirow}
\usepackage{wrapfig}
\usepackage{epsfig}
\usepackage{graphicx}
\usepackage{color}
\usepackage[table]{xcolor}
\usepackage{epsfig}
\usepackage{graphicx}
\usepackage{amsmath}
\usepackage{amssymb,siunitx,subcaption}
\usepackage{footmisc}
\usepackage[pagebackref=true,breaklinks=true,letterpaper=true,colorlinks,bookmarks=false]{hyperref}

\begin{document}

\title{Fine Detailed Texture Learning for 3D Meshes with Generative Models}

\author{
Aysegul Dundar, Jun Gao,  Andrew Tao, Bryan Catanzaro

\IEEEcompsocitemizethanks{
\IEEEcompsocthanksitem  A. Dundar, J. Gao, A. Tao,  B. Catanzaro are  with NVIDIA, CA, USA.
\IEEEcompsocthanksitem A. Dundar is with Department of Computer Science, Bilkent University, Ankara, Turkey
\IEEEcompsocthanksitem J. Gao is with Department of Computer Science, University of Toronto, Canada.
}
}


\IEEEtitleabstractindextext{
\begin{abstract}
This paper presents a method to reconstruct high-quality textured 3D models from both multi-view and single-view images.  The reconstruction is posed as an adaptation problem and is done progressively where in the first stage, we focus on learning accurate geometry, whereas in the second stage, we focus on learning the texture with a generative adversarial network.
In the generative learning pipeline, we propose two improvements.
First, since the learned textures should be spatially aligned, we propose an attention mechanism that relies on the learnable positions of pixels. Secondly, since discriminator receives aligned texture maps, we augment its input with a learnable  embedding which improves the feedback to the generator.
We achieve significant improvements on multi-view sequences from Tripod dataset as well as on single-view image datasets, Pascal 3D+ and CUB.
We demonstrate that our method achieves superior 3D textured models compared to the previous works. Please visit our web-page for 3D visuals: \href{https://nv-adlr.github.io/textured-3d-learning}{https://nv-adlr.github.io/textured-3d-learning}.
\end{abstract}

}

\maketitle

\IEEEdisplaynontitleabstractindextext
\IEEEpeerreviewmaketitle

\section{Introduction}

With GAN based models achieving realistic image synthesis on various objects ~\cite{karras2019style, karras2020analyzing, yu2021dual}, there has been an increased interest to deploy them for gaming, robotics, architectural designs, and AR/VR applications.
However, such applications also require full contollability on the synthesis as well as on the viewpoint.
While recent generative models of images are shown to learn an implicit 3D representation with latent codes that can be manipulated to change the viewpoint of a scene~\cite{karras2019style}, the learned latent space is not disentangled enough to have large changes in the viewpoint with identities of objects remain the same.
Differentiable renderers offer an attractive solution to combine the capabilities of generative models and their abilities to learn from images, and the controllability of viewpoint of the renderers  \cite{chen2019learning, liu2019soft, ravi2020accelerating}.

Various methods are proposed to learn 3D attributes, geometry (triangular meshes) and texture maps from single-view images with differentiable renderers \cite{chen2019learning, liu2019soft, pavllo2020convolutional}.
They deploy an encoder-decoder pipeline powered with deep neural networks and use the image reconstruction loss to train the network weights.
However, with single view images, neural networks only receive supervision from the visible projections of the objects.
Therefore, these networks have no guidance to in-paint the missing regions with consistent texture and mesh predictions.
These algorithms, when trained to predict 3D shapes and texture maps from single images, can learn to reconstruct objects very faithfully from the same viewpoint.
However, from other views, the rendered outputs tend to look unrealistic.

For a faithful recovery of 3D models, current state-of-the-art (SOTA) methods rely on 3D annotations. However, due the difficulty in collecting such annotations, efforts have been limited to few objects such as faces \cite{gecer2019ganfit} and human bodies \cite{zhang2019predicting, lattas2020avatarme}. Finely curated multi-view datasets are the other alternatives which are easier to collect but difficult to annotate for their camera parameters. 
For that reason, synthetic images are used so that models can learn multi-view consistency \cite{chang2015shapenet, choy20163d, chen2019learning}.
However, these models fall short in their ability to recover the 3D properties of real images due to the domain gap between synthetic and real images.
To tackle this problem, Zhang et al.~\cite{zhang2020image} uses a generative adversarial network, StyleGAN \cite{karras2019style}, to generate realistic multi-view datasets.
The latent codes of the network are controlled to generate consistent objects from different view points.
While large improvements are observed with this dataset, we hypothesize that better results can be achieved by learning from real images since these generative models do not have strict disentanglement of shape, texture, and camera parameters.
Therefore the generated multi-view dataset is not perfect especially in presenting the realistic details across views.

\begin{figure*}
\centering
\includegraphics[width=\textwidth]{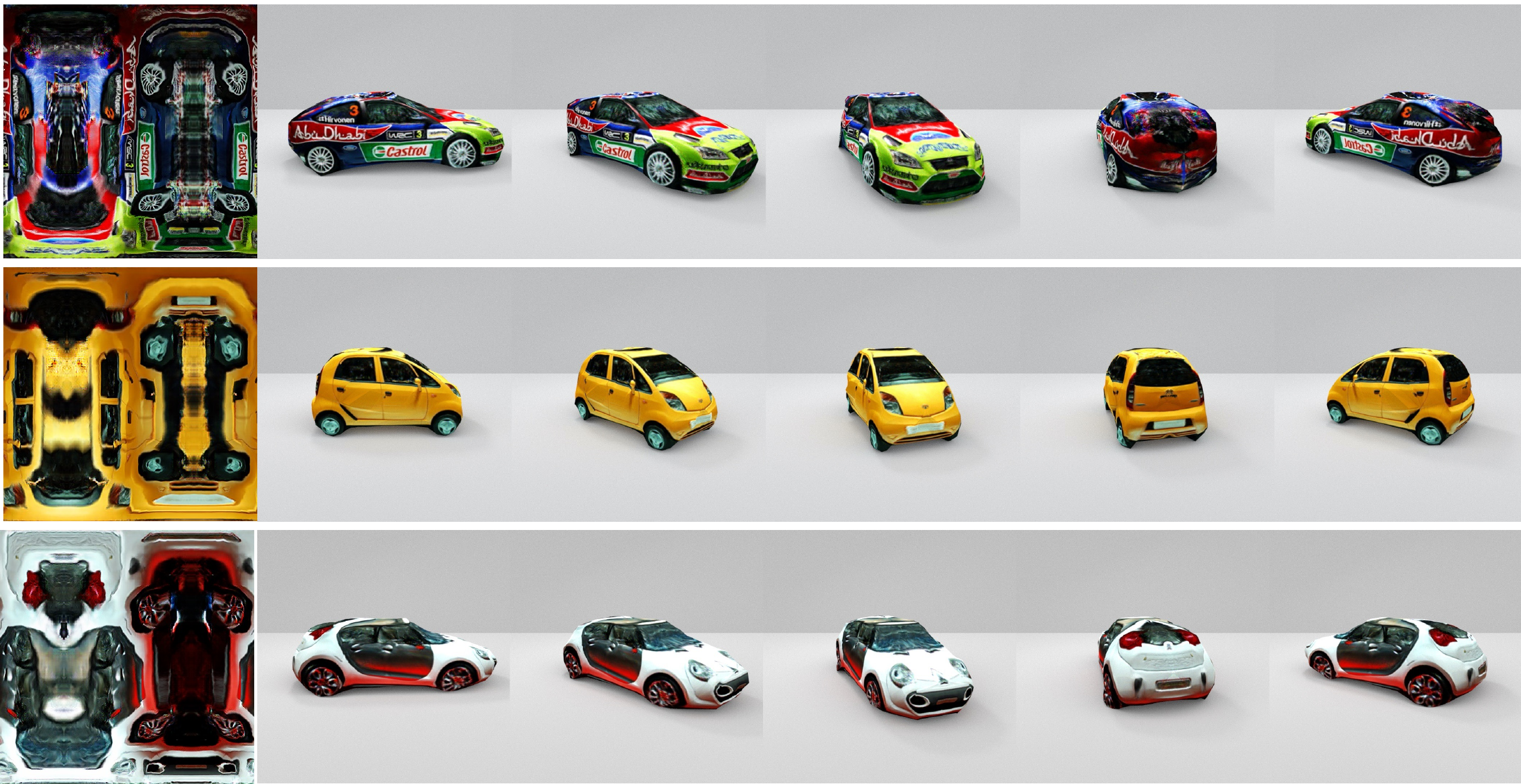}
\caption{Given a multi-view image sequence with no annotations, we infer high quality textured 3D models by utilizing pre-trained networks. The 3D model is presented as triangular mesh and a texture map which can then be used to render novel views.
The texture maps are learned with a novel GAN architecture and present fine details that make textured 3D models look highly realistic.}
\label{fig:teaser}
\end{figure*}

In this work, we focus on inferring realistic textured 3D models -- meshes and texture maps from a collection of multi-view real images with no annotations.
These images are easy to collect as one can record an object by rotating the camera around it.
We do not require the expensive annotation labor of these images such as 2D keypoints, camera pose, or 3D mesh of each frame.
Instead, we rely on the noisy estimate of instance masks and again noisy estimate of camera parameters from pre-trained deep neural network models.
Specifically, we pose the reconstruction of consistent textured 3D models as an adaptation problem which starts from a category-specific 3D reconstruction model trained on a collection of single-view images of the same category.
Our framework first adapts the pre-trained model to learn consistent texture and 3D meshes from the multi-view images, and corrects the camera parameters of initial estimates jointly.
Next, similar to \cite{pavllo2020convolutional}, a 2D convolutional GAN in
UV space for texture map generation is trained.
This final model adds realistic details to the textures and handles variances of the images from the same sequence.
In our GAN pipeline, we propose two improvements one in the generator and one in discriminator where we additionally benefit from learnable position encodings and attention mechanisms. 


Our method synthesizes superior results with a coherent geometry and convincing texture prediction.
To the best of our knowledge, our results are the first ones that achieve realistic synthesis of challenging objects with a differentiable renderer and mesh representation while the model is solely trained on images.
We additionally show that our improvements on the GAN pipeline are not limited to multi-view sequences but also result in better qualitative and quantitative results on the popular single-view image collections, Pascal 3D and CUB datasets.
In summary, our main contributions are:
\begin{itemize}[leftmargin=*]
    \item A framework to synthesize high-quality textured 3D models from multi-view images with no annotations.
    Our framework utilizes pre-trained networks which provide noisy estimates of camera parameters and ground-truth silhouettes.
    Our framework is robust to noisy predictions of the camera parameters and silhouettes and in addition improves those initial predictions. 
    \item A generator and discriminator architecture to handle the unique dataset set-up we have in order to synthesize realistic texture maps.
    Texture maps follow the organization of uv mapping and so they are expected to be spatially aligned across examples (e.g. tires appear in the same place on uv mapping). 
    In our generator and discriminator architectures, we take advantage of this characteristics of texture maps.
    \item High-fidelity textured 3D model synthesis both qualitatively and quantitatively and both on multi-view and single-view datasets. Example generations are shown in Fig. \ref{fig:teaser}.
    More results can be found on the project web-page.
\end{itemize}

\section{Related Work}
\subsection{Generative Adversarial Networks (GANs)}
GAN~\cite{goodfellow2014generative} based models achieve high quality synthesis of various objects which are quite indistinguishable from real images ~\cite{karras2019style, zhang2019self, karras2020analyzing, yu2021dual}.
Controllability of such synthesis is a must for this technology to be incorporated in various applications such as gaming, simulators, and virtual reality.
To achieve controllability, image generation task has been conditioned on different type of inputs such as image patches ~\cite{ zhu2017unpaired, liu2017unsupervised, Zhao2018layout, mardani2020neural},  semantic and instance maps~\cite{isola2017image, wang2018high, park2019semantic, dundar2020panoptic}, and landmarks~\cite{lorenz2019unsupervised, wang2019few, dundar2020unsupervised}.
Image disentanglement for content and style \cite{huang2018multimodal, lorenz2019unsupervised,dundar2020unsupervised, jeon2020cross} in an unsupervised way shows control over pose during inference. 
However, such disentanglement is limited to 2D space and generated images have differences in identity as the pose is manipulated.
Style-based GAN models~\cite{karras2019style, karras2020analyzing} are shown to learn an implicit 3D knowledge of objects without a supervision.
One can control the viewpoint of the synthesized object by its latent codes.
However, in these models, the disentanglement of 3D shape and appearance is not strict and therefore the appearance of objects change as the viewpoint is manipulated.
Recently, 3D-aware generative models are proposed 
\cite{nguyen2019hologan, niemeyer2021giraffe, chan2021pi, gu2021stylenerf} with impressive results but they also do not guarantee strict 3D consistency. 

\subsection{3D Object Reconstruction from Single-view Images}
Learning textured 3D models from images has become possible with differentiable renderers~\cite{loper2014opendr, kato2018neural, liu2019soft, chen2019learning, ravi2020accelerating}.
These renderers can be coupled with deep neural networks and  from input images deep networks can be trained to predict 3D mesh representations, texture, and lighting parameters. From these 3D attributes the renderers can  output images with any camera views.
The image reconstruction losses between the input image and the rendered image by the input view can be back-propogated via differentiable renderers to train the deep neural network parameters ~\cite{kanazawa2018learning, chen2019learning, goel2020shape, li2020self, henderson2020leveraging}.
However, inferring these 3D attributes from  single 2D images is inherently ill-posed problem.
Especially without ground-truth camera parameters, learning shape and viewpoint in an unsupervised way has only shown to be possible recently \cite{goel2020shape} but the results are still significantly in lower quality than its counterparts which use ground truth camera parameters.
Even with the known camera parameters, learning 3D inference from single 2D images is still ill-defined given that the invisible mesh and texture predictions receive no gradients during training ~\cite{kanazawa2018learning, chen2019learning, goel2020shape, li2020self, henderson2020leveraging}.
These algorithms while can recover
3D shape and texture map from a single image that can reconstruct the object very faithfully from the same viewpoint, when viewed from a different point, the rendered outputs tend to look unrealistic.

\subsection{3D Object Reconstruction from Multi-view Images} Multi-view image datasets provide a solution for the limited supervision problem of single-view image datasets.
However, due to the expensive annotation of 
multi-view image datasets for their 2D keypoints or camera pose, they are small in scale.
Recently, these datasets have been tremendously explored with a method called Neural Radiance Fields (Nerf) to explore implicit geometry \cite{mildenhall2020nerf, martin2021nerf, wang2021nerf}.
These models overfit to a sequence and can not be used to train on single-view image collections.
In our work, we are interested in mesh representations due its efficiency in rendering.
Additionally, we are interested in a framework that can learn both from single-view and multi-view image collections.
To infer mesh representations, multi-view datasets are also shown to be beneficial based on the experiments with synthetic datasets~\cite{chang2015shapenet, tulsiani2017multi, tulsiani2018multi}.
However, those results do not translate well to real image inferences because of the domain gap between synthetic and real images.
To generate a realistic multi-view dataset with cheap labor cost, Zhang et. al.~\cite{zhang2020image} use a generative adversarial network by controlling the latent codes and generate coarsely consistent objects from different view points.
We hypothesize that better results can be obtained from real images since the generative model used by Zhang et. al.~\cite{zhang2020image}  do not have strict disentanglement, and therefore the generated multi-view dataset is not perfect especially in presenting the realistic details across views.
In another work, recovering animal geometries is set as an online adaptation problem from videos collected  in the wild \cite{li2020online}.
That work allows as rigid as possible changes across frames from a base shape which is constructed by applying K-Means clustering to all
meshes reconstructed by CMR \cite{kanazawa2018learning}.
In our work, we do not require base shapes, and are interested in 360\degree rotation of objects to model the complete texture.
We are further interested in generating high quality texture maps that can compete with the realism quality of recent GAN based methods on 2D image synthesis.


\begin{figure*}
    \centering
    \includegraphics[width=0.95\linewidth]{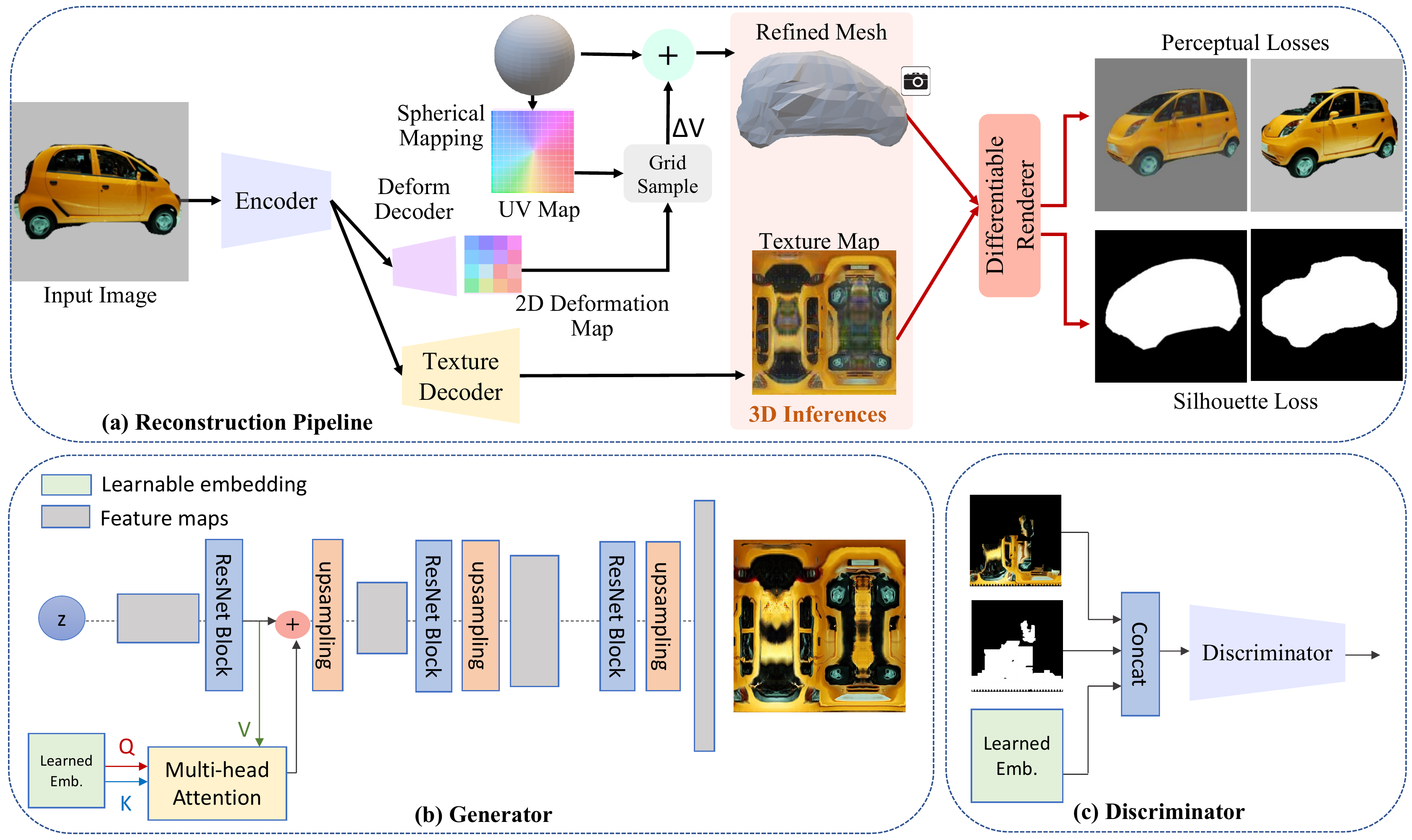}
    \caption{\textbf{Overall Pipeline.} (a) Reconstruction pipeline estimates convolutional mesh and texture representations \cite{pavllo2020convolutional}.
    Different than \cite{pavllo2020convolutional}, we reconstruct this car from a different view with a differentiable renderer.
    The perceptual loss is calculated between the reconstructed image and ground truth image and a silhouette loss is calculated between the rendered silhouette and ground-truth mask to train the networks.
    The camera parameters of the rendered image are optimized jointly which is removed from the Figure for brevity.
    Next we learn fine detailed textures with GAN pipeline.
     In GAN training pipeline, we propose two improvements. The improvements are tailored to the unique dataset set-up we have where images, texture maps, are spatially aligned for each sample. (b) Since the textures are always spatially aligned, in the Generator, we design an attention mechanism to learn which positions in the texture map should be considered together. 
     (c) In the Discriminator, with the same intuition, we learn a positional embedding so that the patch discriminator specializes for different parts.
    }
    \label{fig:over_method}
\end{figure*}

\section{Method}

For multi-view datasets with no annotations, our method poses the 3D reconstruction as an adaptation problem which starts with a pretrained network that encodes shape, texture, and camera parameters of a given image.
We start fine-tuning this network on multi-view image datasets with the initial estimates of the camera parameters from the pretrained network.
The learning is separated into two pipelines.
First, we focus on learning accurate geometry, whereas in the second stage, we focus on learning high quality texture maps with the addition of a generator and GAN loss.
For single-view datasets, we experiment on datasets that have camera view annotations.
We again have the first stage where we focus on learning accurate geometry by using the ground-truth camera parameters and the second stage to learn high quality texture maps.
Our improvements on learning high quality texture maps also apply to single-view image datasets and improve the results.

\subsection{Mesh Learning}
We follow the same encoder-decoder architecture from Pavllo et. al ~\cite{pavllo2020convolutional} that encodes images into 3D geometry and textures which are fed into a differentiable renderer as shown in Fig.~\ref{fig:over_method}.
In this set-up, we are only interested in learning consistent 3D geometry across views but we also learn texture maps to enable the learning with image reconstruction loss.
We use DIB-R~\cite{chen2019learning} as our differentiable renderer. 
For the camera model, we adopt a weak-perspective model where the pose of an image is described by  rotation $q \in \mathbb{R}^4$, scale $s \in \mathbb{R}$, and  translation $t \in \mathbb{R}^2$ matrixes.
We use the initial estimates from the pretrained model and additionally learn the offset correction parameters for $q$, $s$, and $t$. 
We adapt the pretrained model on a set of images that belong to the same object.
In the encoder-decoder pipeline, we feed an input image to the encoder and ask the framework to reconstruct the image from a different view as shown in Fig.~\ref{fig:over_method}.
This way the estimates are not biased by the input image.
For a given training sample, we load two randomly selected camera views and optimize the camera parameters for the target view given the state of the geometry and texture.
In our trainings with single-view image datasets, we directly optimize the model to reconstruct the input view.

 {\noindent \bf{Losses.}} We use similar losses as \cite{kanazawa2018learning, chen2019learning, pavllo2020convolutional} to train our deep neural networks in the reconstruction pipeline.
We use $\mathcal{L}_{percp}$ perceptual loss between the input image ($I$) and the rendered image ($I_r$)  from Alexnet ($\Phi$) at different feature layers ($j$) between these images. 
We mask input image as well to remove the background, therefore perceptual loss is only calculated among the objects.
\begin{equation}
\label{eq:percep}
\small
    \mathcal{L}_{percp} = ||\Phi_{j}(I) - \Phi_j(I_r) ||_2
\end{equation}
We additionally use a silhoutte loss which calculates intersection-over-union between the silhouette rendered ($S_r$) and the silhouette ($S$) (instance mask segmentation) of the input object.
\vspace{-1pt}
\begin{equation}
\label{eq:sil}
\vspace{-1pt}
\small
\mathcal{L}_{sil} = 1 - \frac{||S\odot S_r ||_1}{||S+S_r - S\odot S_r ||} 
\end{equation}
Similar to~\cite{chen2019learning, liu2019soft, pavllo2020convolutional}, we also regularize predicted mesh using a smoothness loss  ($\mathcal{L}_{sm}$) constraining neighboring mesh triangles to have normals with similar directions.
Similar to~\cite{kanazawa2018learning}, we learn to predict camera pose 
using a simple $\mathcal{L}_2$ regression loss ($\mathcal{L}_{cam}$) between ground truth and predictions that we get from an additional Resnet18 backbone that also takes images as input and outputs rotation, scale and translation parameters.
Following are our base losses:
\vspace{-2pt}
\begin{equation}
\label{eq:total_loss}
\begin{multlined}
\small
\mathcal{L}_{total} = 
 \lambda_p \mathcal{L}_{percp} + \lambda_s \mathcal{L}_{sil}   + \lambda_c \mathcal{L}_{cam} +   \lambda_{sm} \mathcal{L}_{sm}   
\end{multlined}
\end{equation}

Our $\lambda$ coefficients are:  $\lambda_p=1$, $\lambda_s=1$,  $\lambda_c=1$, $\lambda_{sm}=0.00005$. The loss from Eq. \ref{eq:total_loss} is used to train the the models on Pascal 3D+ dataset and CUB dataset. In the adaptation trainings on Tripod dataset, we use the same loss and parameters except the camera loss $\mathcal{L}_{cam}$ since we do not have ground-truth camera parameters.

\subsection{Fine Detailed Texture Learning}
After learning the mesh representation, we use generative learning pipeline \cite{pavllo2020convolutional} to convert texture learning into 2D image synthesis task.
The input images are projected onto the UV map of the predicted mesh template based on the optimized camera parameters of each image.
This projection outputs pseudo ground-truth texture maps with inverse rendering.
In this process, mesh predictions are transformed onto 2D screen by projection with optimized camera parameters.
Then transformed mesh coordinates and UV map coordinates are used in reverse way and real images are projected onto UV map with the renderer.
Visibility masks are also obtained in this set-up which are used to mask the projected textures and also used during GAN based training to mask the losses as well as generated textures to prevent a distribution mismatch between the generated and pseudo ground-truth texture maps.

The accuracy of obtaining consistent texture pseudo ground-truth depends on the accuracy of predicted meshes and camera parameters.
If either one of them is wrong, the meshes will project to a wrong placement on the real image and wrong pixels will end up in the pseudo ground-truth texture map.
Therefore, the corrections we make in the camera parameters by optimization become crucial in this step.
We find the second important aspect of the ground-truth generation to be the alignment in the borders.
If they are not correctly aligned, the masked out gray regions can leak into our texture maps as visible parts.
Since the silhouette masks are generated by Mask-RCNN, we find them to be not precisely contouring the object borders on Tripod dataset.
An example of such imprecise mask can be seen in Fig. \ref{fig:over_method} shown for the silhouette loss calculation. The Mask-RCNN generated masks and correspondingly the input images may have some impurities in the borders. 
On the other hand, we observe that the masks generated by 3D mesh projection of our network are more accurate since the reconstruction pipeline is trained with multi-view consistency. 
Therefore, when pseudo ground-truth texture maps are generated, we use the silhouettes estimated by the 3D mesh projection to mask the inputs. 

With this pseudo ground-truth texture maps, shown as input to the Discriminator in Fig. \ref{fig:over_method}, a GAN based model is trained for each sequence on Tripod dataset and on single-view image datasets Pascal 3D+ and CUB.
The GAN based training is only performed for texture generation. 


\subsubsection{Position-based Attention in Generator} The generated textures are expected to show no spatial variance as they have to be aligned with their uv map and corresponding 3D mesh vertices (e.g. tires always appear at the same position). We take advantage of this unique set-up of generated data and propose a learnable position attention in the generator as shown in Fig. \ref{fig:over_method}(b).
The key and the query of the attention module are learnable embeddings.
Key is used to correlate with the query to output the attention maps.
The attention maps model which pixels should be attended when synthesizing new features based on their spatial locations.
The value is the features generated in the previous layer.

The attention is applied as follow:
\begin{equation}
\begin{split}
\bar{\mathbf{F}}^\textit{attn} & \doteq\mathsf{attn}\left(\mathbf{K}(\mathbf{P}),\mathbf{Q}(\mathbf{P}),\mathbf{V}(\mathbf{F})\right)\\
\mathbf{F}^\textit{out} & =\bar{\mathbf{F}}^\textit{attn}+\mathbf{F}\\
\end{split}
\label{eq:output_self_attn}
\end{equation}
where $\mathsf{attn}$ is the multi-head attention \cite{vaswani2017attention}, $\mathbf{P}$ is the learnable embedding, and $\mathbf{F}$ is the feature maps.
$\mathbf{K}$, $\mathbf{Q}$, and $\mathbf{V}$ are convolutional layers.
$\mathbf{K}$ and $\mathbf{Q}$ are used to find good representations during training.
During inference, attention maps can be pre-computed.
The attention layer is added to the early layer of the generator. 
A second attention layer did not show a benefit which is in agreement with other GAN works with attention \cite{zhang2019self, yu2021dual}.
Additional details of the network architecture is given in Sec. \ref{sec:training}.

\subsubsection{Learnable Position Embedding in Discriminator}
Discriminator is guiding the generator to synthesize a complete texture map while only seeing partially erased textures as shown in Fig. \ref{fig:over_method}(c).
This presents a unique challenge to the generation process.
Even though the visibility mask is also fed to the discriminator and so the discriminator can learn to separate what is visible and not, the outputs of the discriminator is still affected by the black pixels in the input.
This prevents us to use a discriminator with a large receptive field as black pixels start to leak in to the generated images.
A multi-scale discriminator with a receptive field of $16\times16$ works reasonable well, but due to the limited receptive field, it does not have the capacity to realize which patch in the input image it is classifying.
Since the presented texture maps to the discriminator are supposedly always be spatially aligned, the discriminator can benefit from a learnable positional embedding.
With this motivation, we also concatenate a learnable embedding with the same height and width of the texture maps to the input of the disciminator instead of a fixed position embedding.
This modification results in significant improvements on the results.

\section{Experiments}
\label{sec:exp}

\begin{figure}
\centering
\scalebox{1.0}{
\addtolength{\tabcolsep}{-5pt}   
\begin{tabular}{cccc}

\includegraphics[width=0.15\textwidth]{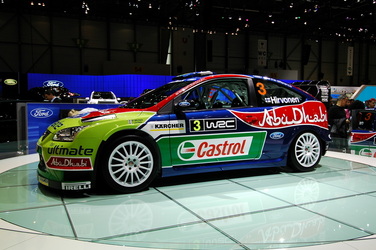} &
\includegraphics[width=0.15\textwidth]{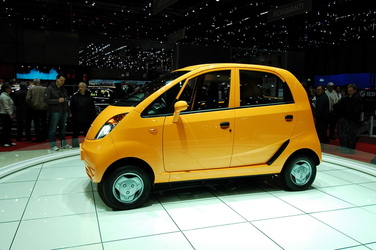} & 
\includegraphics[width=0.15\textwidth]{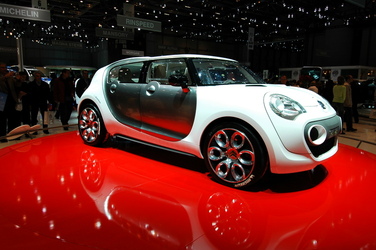} \\
\includegraphics[width=0.15\textwidth]{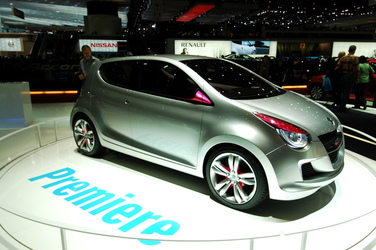} &
\includegraphics[width=0.15\textwidth]{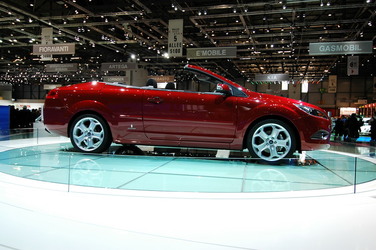} &
\includegraphics[width=0.15\textwidth]{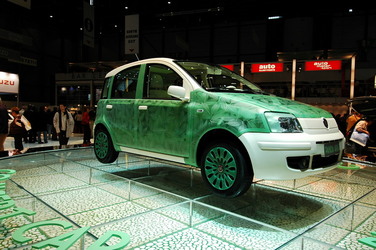} \\
\end{tabular}
}
\caption{Example images from six sequences of Tripod dataset.}
\label{fig:dataset}
\end{figure}

\subsection{Datasets and Preprocessing}

\subsubsection{Multi-view Datasets.}
We run experiments on multi-view Tripod dataset \cite{ozuysal2009pose}.
This dataset contains diverse car sequences on a rotating platform as they rotate by 360 degrees.
Each sequence contains around 90 to 120 images.
There is one image approximately every 3-4 degrees and we take advantage of this characteristic when detecting the images with wrongly optimized camera parameters and remove those images in our preprocessing.
It does not have any annotations and we generate the silhouette masks with Mask-RCNN \cite{he2017mask}.
This dataset exhibits a domain gap with the COCO dataset \cite{lin2014microsoft} (which Mask-RCNN is trained on) as it is collected on a shiny platform in a showroom as shown in Fig. \ref{fig:dataset}.
Due to this domain gap, we find Mask-RCNN to be not as accurate on Tripod dataset as it is on COCO dataset but due to the multi-view image set-up in the reconstruction pipeline, the errors on the mask silhouettes do not propagate to the mesh learning.

We also predict camera parameters of Tripod images with a network trained on Pascal 3D+ dataset \cite{xiang2014beyond} car category.
These predicted camera parameters are again very noisy due to the domain gap between Tripod and Pascal 3D+ datasets. 
Our framework is also robust to the errors in camera parameters and optimize them jointly with multi-view consistency loss.
Alternatively, to estimate camera parameters, we also run COLMAP algorithm \cite{schoenberger2016mvs}.
However, bundle adjustment did not converge on these challenging sequences with
images that have many high lights which are not consistent among different views.

We optimize the deep network parameters and the camera parameters of the image that we use as the target. 
We find that with this joint optimization, the initial camera parameters are mostly corrected. However, some challenging views start from a very bad estimate of the pretrained network.
For example, instead of estimating the camera facing front side of the car, the network estimates the camera view point as it is facing the back side of the car.
The camera parameters do not get corrected from that state.
These images can be identified easily in the multi-view dataset as the images in a sequence are taken of a smoothly rotating platform.
Therefore, when the predicted camera parameters have large differences in consecutive frames, they belong to the examples with wrongly estimated camera parameters.
We remove them from the dataset to obtain our pruned dataset and train the network again.
Removing these images provide with additional improvements on the texture and 3D mesh.
We also tried updating camera parameters with multiple camera hypothesis \cite{goel2020shape} but did not see these parameters get corrected for those wrongly estimated images.
Since we have a collection of them available, we choose to remove these images. For other settings, it may be important to invest more in correcting the camera parameters for all images but it is not found crucial in this set-up.

\begin{figure}
\centering
\scalebox{1.0}{
\addtolength{\tabcolsep}{-5pt}   
\begin{tabular}{ccc}

%

 &
\includegraphics[width=0.16\linewidth]{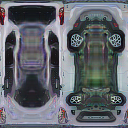} &
\includegraphics[width=0.80\linewidth]{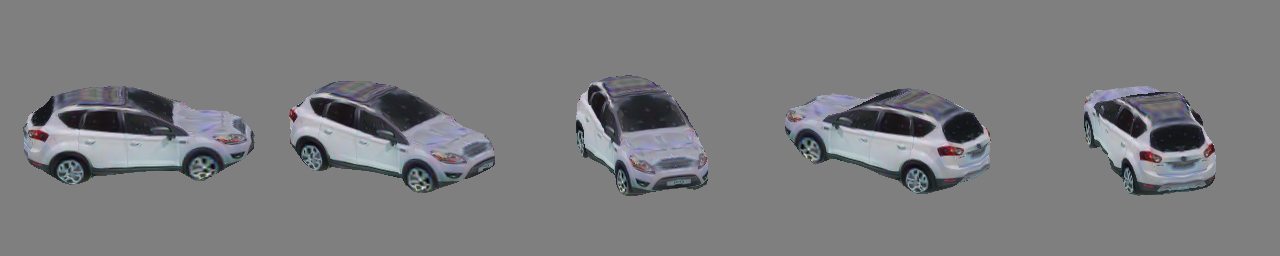} \\
 &
\includegraphics[width=0.16\linewidth]{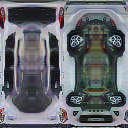} &
\includegraphics[width=0.80\linewidth]{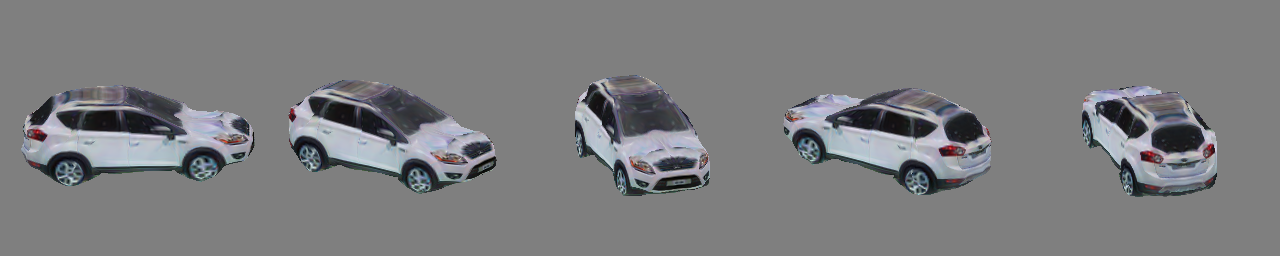} \\

\end{tabular}
}
\caption{3D reconstruction outputs of models trained with unpruned and pruned datasets. First column shows texture map predictions, other columns show the rendering results from different viewpoints. First row, mesh learning results from the multi-view dataset. Second row, mesh learning results after removing the frames with wrong camera estimation. Notice the improvements on the headlight and plate after pruning the dataset.}
\label{fig:ablation_results}
\end{figure}

Adapting the network on the pruned dataset,  improves the 3D model as shown in Fig. \ref{fig:ablation_results}.
The improvement is especially visible around the  headlight and plate areas in texture maps as well as on the shape.
The corrections we make in the camera parameters by optimization and pruning the dataset of the wrongly camera predicted images also become crucial in the pseudo ground-truth generation step.
Our pre-processing set-up of removing frames with wrong camera estimations also prevent those wrong pseudo ground-truth texture maps occurrence.
Pruning the dataset improves the results.
In our ablation study as shown in Table \ref{tab:sup_abl}, we compare GAN models trained on pruned (first row) and unpruned datasets (second row).
Removing these images gives us a large improvement in FID. 

\begin{table}
    \centering
    \begin{tabular}{l|c|c}
       \toprule
    \multicolumn{3}{c}{Tripod Dataset \cite{ozuysal2009pose}}\\
    \toprule
    
         Method &  FID & $ \Delta $ \\
         \hline
         GAN model - Baseline & 82.19 & 0 \\
         GAN model - Unpruned dataset & 100.43 & +19.17 \\
        \hline 
    \end{tabular}
    \caption{FID scores in Tripod dataset. We show the FID scores of models trained on pruned (baseline) and unpruned datasets.
    Results are averaged for 19 sequences.
    Lower is better; best result is shown in bold.}
    \label{tab:sup_abl}
\end{table}

\subsubsection{Single-view Datasets}
We also evaluate our method on  Pascal 3D+ dataset \cite{xiang2014beyond} and  CUB \cite{welinder2010caltech} bird dataset.
We use same train-test split as that provided by DIB-R \cite{chen2019learning}.
These datasets provide single-view images with keypoint annotations from which camera pose is approximately estimated by using structure-from-motion \cite{schoenberger2016sfm}. 
Ground-truth masks are available for a subset of them and for the others we generate them with off the shelf instance segmentation model, Mask-RCNN \cite{he2017mask}.
Since the mask and camera parameters are accurate for these datasets,  we directly use them to evaluate our improvements on the generative pipeline.

\begin{figure}
\centering
\includegraphics[width=0.48\textwidth]{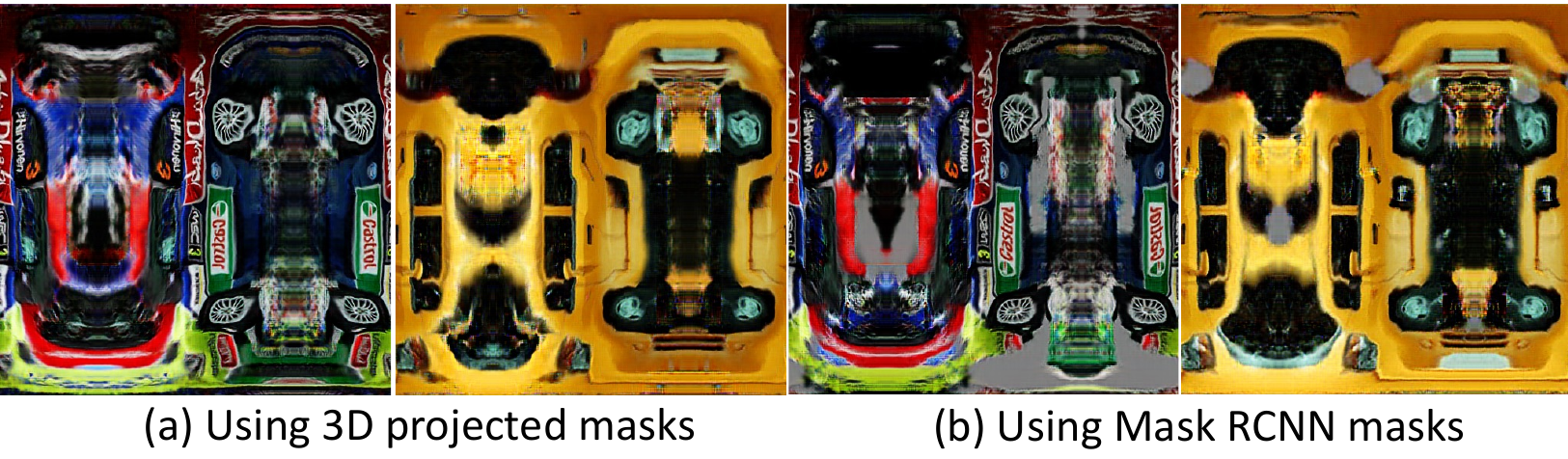} 
\caption{(a) Textures generated by our baseline which is trained with pseudo-ground truth texture maps that uses masks generated by 3D mesh prediction and (b) Textures generated by the network which is trained with pseudo-ground truth texture maps that uses masks obtained by Mask-RCNN. Mistakes on the mask predictions cause a leakage of black pixels into the texture generation. }
\label{fig:abl}
\end{figure}

\begin{figure}
\centering
\scalebox{1.0}{
\addtolength{\tabcolsep}{-5pt}   
\begin{tabular}{cc}

\includegraphics[width=0.16\linewidth]{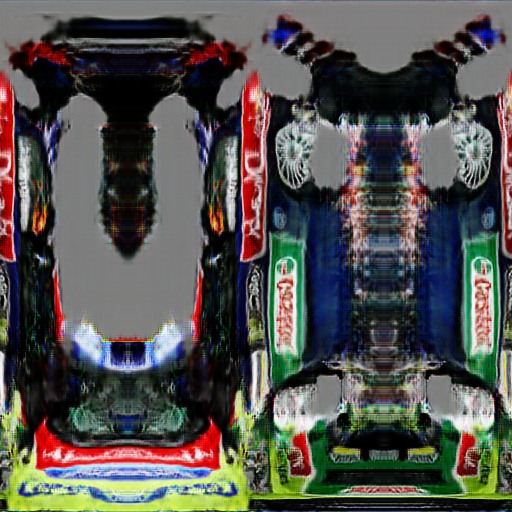} &
\includegraphics[width=0.80\linewidth]{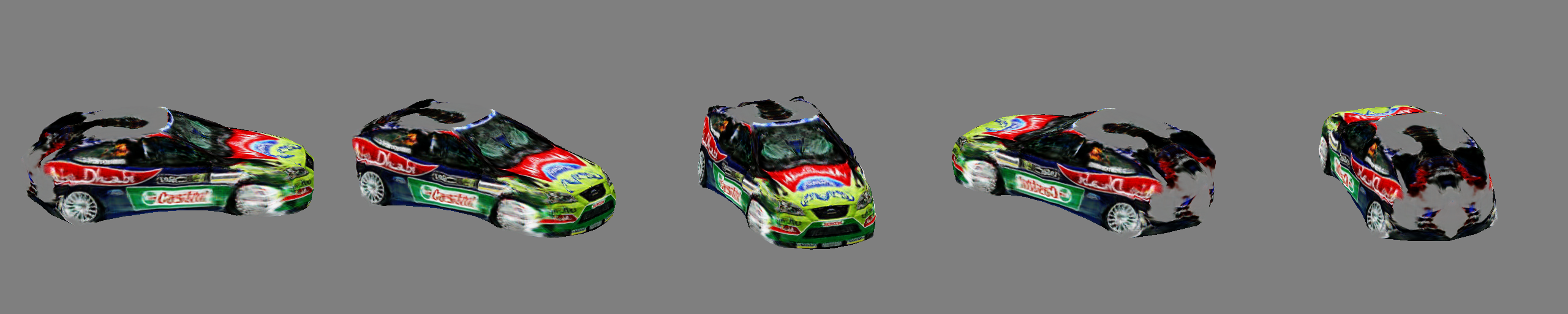} \\
\includegraphics[width=0.16\linewidth]{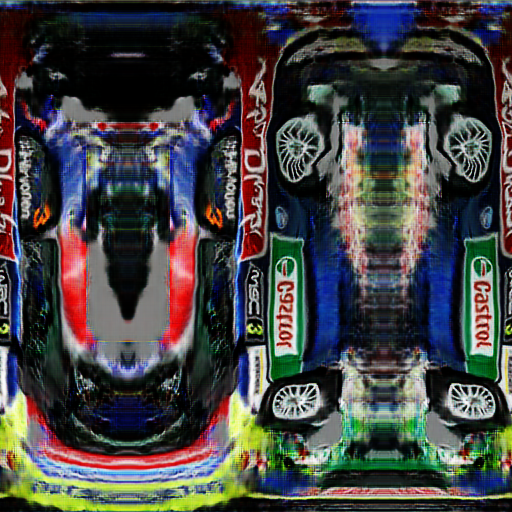} &
\includegraphics[width=0.80\linewidth]{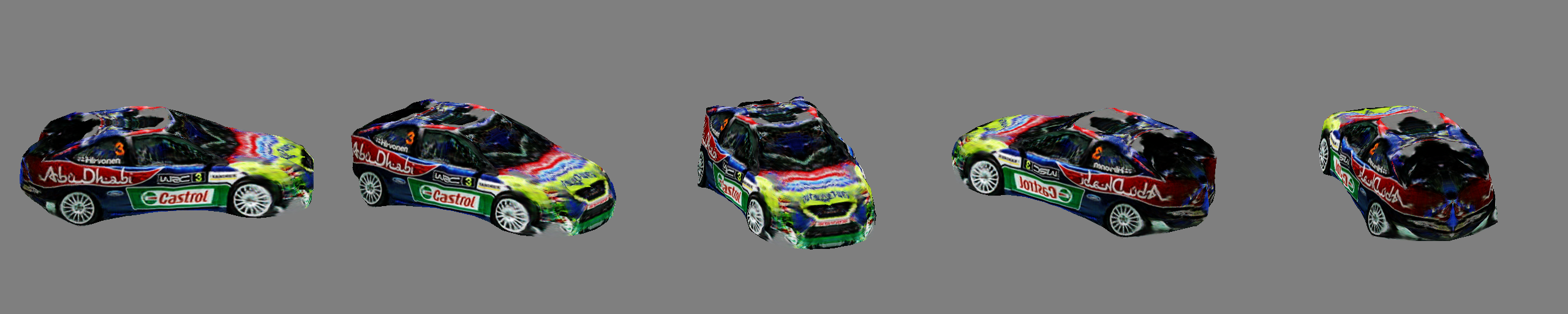} \\

\end{tabular}
}
\caption{Texture and rendering results by the network which is trained with pseudo-ground truth texture maps that uses masks obtained by Mask-RCNN. First column shows texture map generations, and other columns show the rendering results from different viewpoints.
We find the grey pixels always exist on the generated texture maps, sometimes more subtle as in the second row and sometimes more significant as in the first row.}
\label{fig:mask_examples}
\end{figure}

\begin{figure*}
\centering
\includegraphics[width=0.98\textwidth]{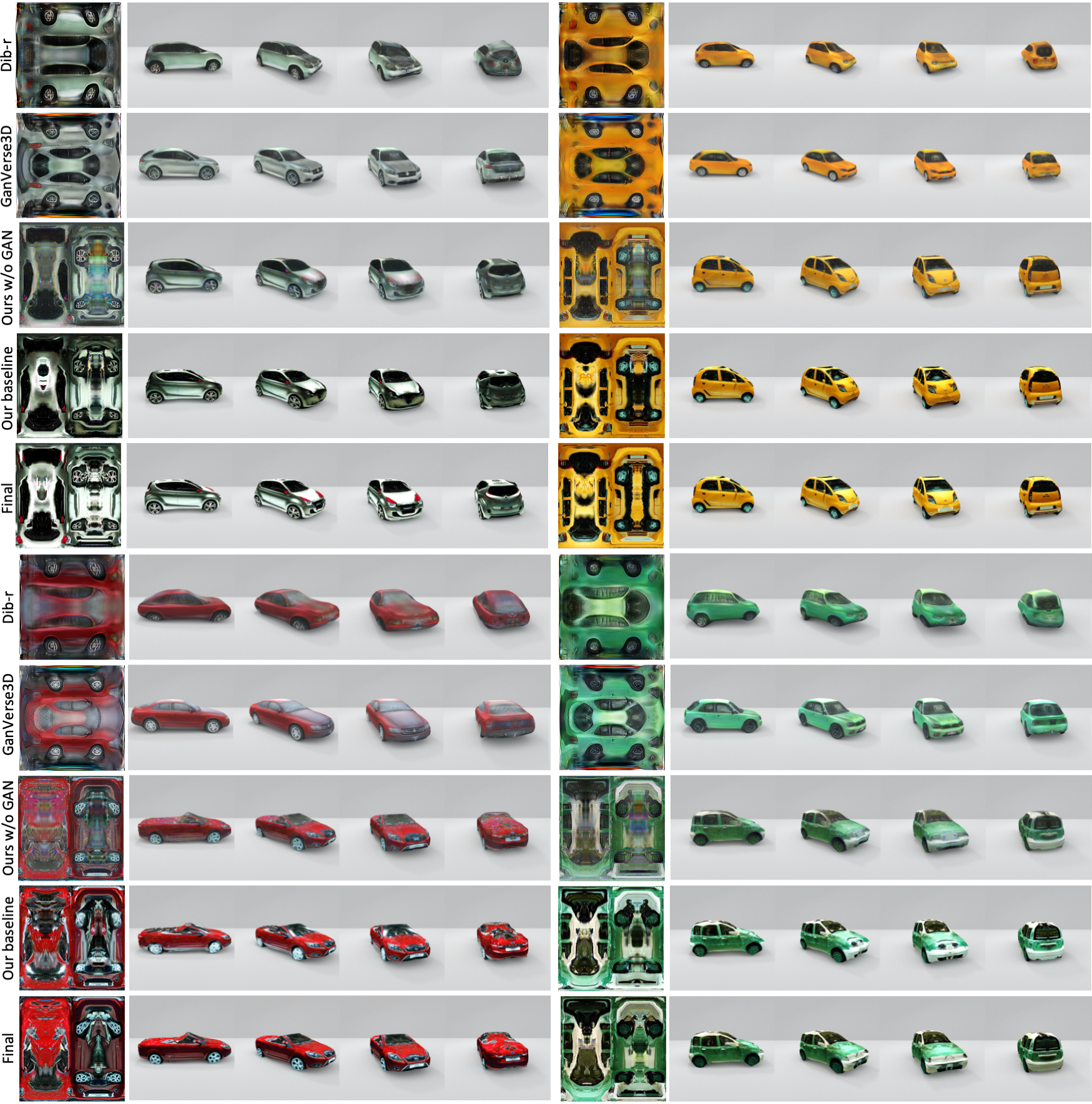} 
\caption{Figure shows texture map generations, and rendering results from different viewpoints.
First row and second row present the results of Dib-r \cite{chen2019learning} and GANVerse3D model \cite{zhang2020image}, respectively. Dib-r model is trained on single view image dataset whereas GANVerse3D model is trained on multi-view images that are generated by StyleGAN.
They are not comparable to our work since we adapt our model to each sequence, however, they still provide important information on the generalization of these models and their ability to render realistic details.
Ours without GAN presents the reconstruction model's results after adapting to each sequence.
It  does not output as crisp images as 2D GAN models.
Our baseline model includes our improvements on the dataset generation pipeline but it uses the network architectures from \cite{pavllo2020convolutional}.
Final model incorporates our network changes in the GAN pipeline. The improvements are significant as was also measured in FID.
Ground-truth images of these models can be seen in Fig. \ref{fig:dataset}.
}
\label{fig:results}
\end{figure*}


\subsection{Evaluation}
We follow the evaluation protocol for 2D image synthesis and report FID scores by rendering images from multiple views  and comparing them against the training data.
Images are rendered for novel views and are scaled to $299\times299$ to calculate FIDs.
On Tripod dataset, we calculate the best FID scores for each sequence and average them for over 19 sequences.
On Pascal 3D+ and CUB datasets, we train each model 5 times with different random seeds and report the average FIDs of these runs.

\subsection{Model and Training Details}
\label{sec:training}

We have two pipelines as proposed by \cite{pavllo2020convolutional} and we build our experiments on the author's released code-base. First pipeline, the reconstruction network where we learn our final mesh results and optimized camera parameters, is shown in Fig.~\ref{fig:over_method}. We have a shared feature extractor for predicting deformation and texture maps. We also have an additional Resnet18 network \cite{he2016deep} pre-trained on ImageNet that is used to predict camera poses which is omitted from this figure. We predict the camera poses by adding a single layer over ResNet18. This module is only used for the pre-trained network where we have ground-truth camera parameters.
We train an encoder-decoder network on Tripod datasets, Pascal 3D+, and CUB datasets on a single GPU with batch size of 50.
For tripod datasets, we start from a pretrained network that was trained on Pascal 3D+ dataset.
For the pretrained network, the training completes in 18 hours and for the adaptation networks, the training completes in 8 hours.

In the second pipeline, we set a Generator and Discriminator to learn realistic texture maps.
The architecture of the generator starts with a sampled noise from normal distribution with a shape of $b\times64$ where $b$ is the batch size. Batch dimension is dropped for brevity in the rest of the section.
The initial map is fed to a linear layer to output $512\times4\times8$ feature maps.
It goes through 7 blocks of residual layers, ReLU, and batch normalization layers, and 6 upsampling layers in between to output final texture maps with a dimension of $3\times256\times512$.
A reflection symmetry is applied to output $3\times512\times512$ texture maps.
Position based attention is applied in the second layer.
First learnable embedding goes through two different convolutional layers to output query and key with output channel of $32$.
The attention map is calculated after sending the matrix multiplication to softmax operation.
Value also goes through a convolutional layer and multiplied with attention map to output the final results which goes through a batch normalization layer.
The calculated result is added to the input feature maps as residuals.
The discriminator takes the generated and real $3\times512\times512$ texture maps, and the pseudo ground-truth visibility mask.
Generated textures are also multiplied with the masks to prevent a mismatch between the real-fake data distributions.
The discriminator adopts a multi-scale architecture with two scales one operates on $32\times32$ patches, the other $16\times16$.
We concatenate the input with learnable positional embeddings on both scales.
This second set-up is trained on 4 GPUs with batch size of 32 for around 18 hours.

\begin{table}
    \centering
    \begin{tabular}{l|c|c}
       \toprule
    \multicolumn{3}{c}{Tripod Dataset \cite{ozuysal2009pose}}\\
    \toprule
    
         Method &  FID & $ \Delta $ \\
         \hline
         GAN model - Baseline & 82.19 & 0 \\
         GAN model - Using Mask-RCNN masks & 84.90 & +2.71\\
         \hline
         +Discriminator w/ learned emb. & 66.57 & -15.62 \\
         +Generator w/ attention & \textbf{64.55} & -17.64 \\
        \hline 
    \end{tabular}
    \caption{FID scores in Tripod dataset. We show the FID scores of various set-ups and their difference with the baseline in the last column. 
    Results are averaged for 19 sequences.
    Lower is better; best result is shown in bold.}
    \label{tab:abl}
\end{table}

\begin{table}
    \centering
    \begin{tabular}{l|c|c}
       \toprule
        \multicolumn{3}{c}{Pascal 3D+ Dataset \cite{xiang2014beyond}}\\
    \toprule
         Method &  avg. FID & best FID \\
         \hline
         ConvMesh \cite{pavllo2020convolutional} & 29.06 & 27.81\\
         \hline
         +Discriminator w/ learned emb. & 26.95 & 26.18  \\
         +Generator w/ attention & \textbf{26.79} & \textbf{25.63} \\
        \hline 
    \end{tabular}
    \caption{FID scores on Pascal3D car dataset. 
    Each model is trained 5 times with different random seeds.
    The best FID's of these runs are averaged  and best results among them are presented.
    Lower is better; best result is shown in bold.}
    \label{tab:p3d}
\end{table}

\begin{table}[]
    \centering
    \begin{tabular}{l|c|c}
       \toprule
        \multicolumn{3}{c}{CUB Dataset \cite{welinder2010caltech}}\\
    \toprule
         Method &  avg. FID & best FID\\
         \hline
         ConvMesh \cite{pavllo2020convolutional} & 40.80 &  39.21\\
         \hline
         +Discriminator w/ learned emb. & 39.47 & 37.82  \\
         +Generator w/ attention & \textbf{38.45} & \textbf{37.02} \\
        \hline 
    \end{tabular}
    \caption{FID scores on CUB bird dataset. 
    Each model is trained 5 times with different random seeds.
    The best FID's of these runs are averaged  and best results among them are presented.
    Lower is better; best result is shown in bold.}
    \label{tab:cub}
\end{table}

\begin{figure}
\centering
\includegraphics[width=0.48\textwidth]{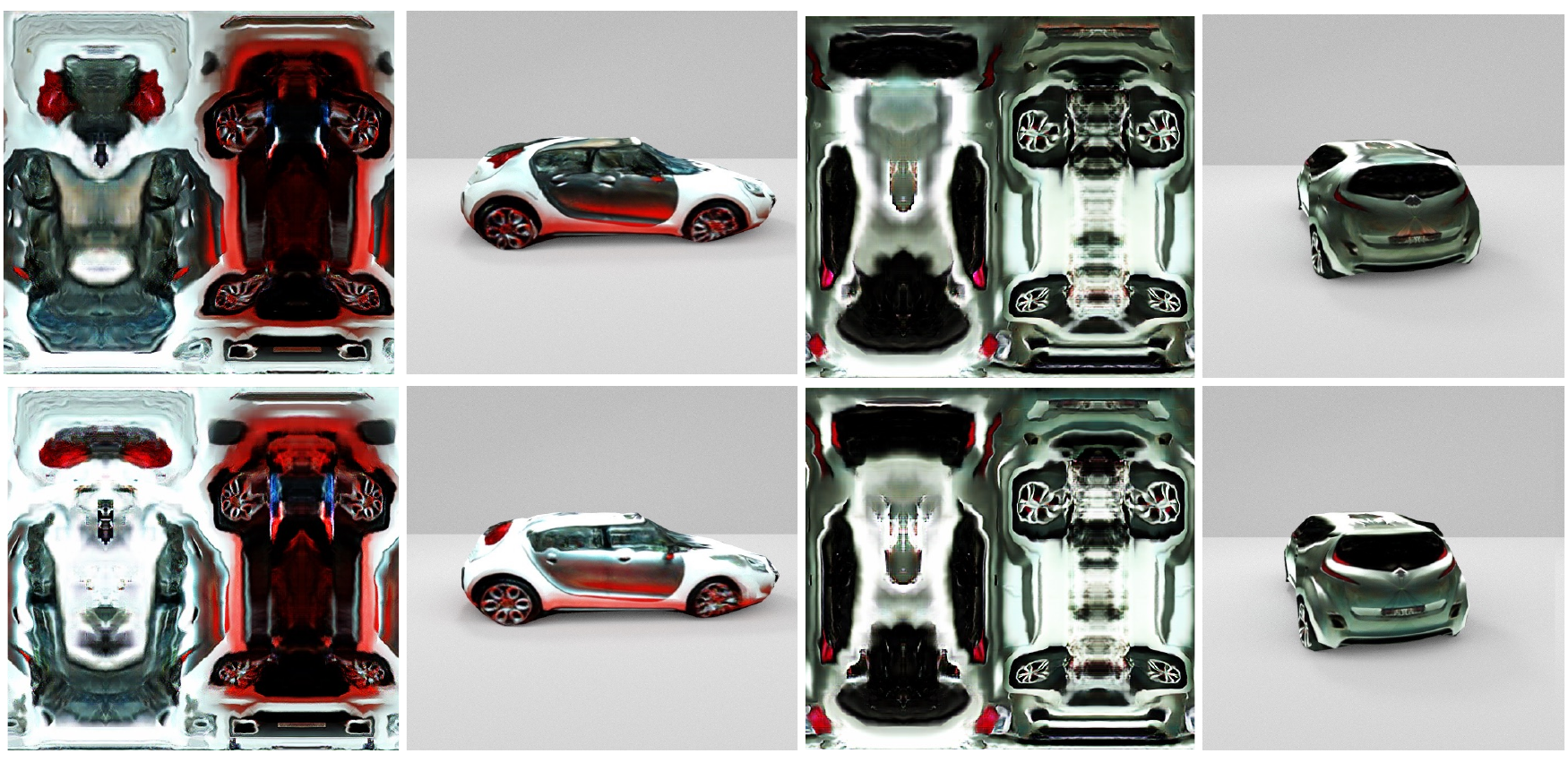} 
\caption{Pseudo ground-truth textures show variety due to lighting and reflection changes from the capturing process because cars are rotating and receiving light from different relative directions.
GAN based training enable us to synthesize diverse texture maps in this context.
The differences in the texture maps model different lighting directions and different reflections on the glasses. }
\label{fig:final_diverse_2}
\end{figure}

\subsection{Quantitative Results}

\subsubsection{Multi-view Datasets}

On Tripod dataset, there is no prior work for 3D textured model reconstruction. We obtain baselines with various set-ups.
Note that, other models that are trained on Pascal 3D+ with single view image reconstruction perform poorly on this dataset, and not comparable to our work \cite{chen2019learning, kanazawa2018learning, pavllo2020convolutional, goel2020shape, bhattad2021view} which is discussed more in depth in Qualitative Results section (Sec. \ref{sec:qual}) and in Fig. \ref{fig:results}.

In our ablation study as shown in Table \ref{tab:abl}, we compare baseline and using Mask-RCNN masks network which stands for the model trained with pseudo-ground truth obtained by first masking the images with Mask-RCNN instance segmentation predictions whereas baseline uses the 3D projected masks from the renderer.
Using Mask-RCNN predictions causes degradation in FID. This happens because of the impurities in the instance segmentation masks and resulting in masked black pixels to end up in pseudo ground-truth texture maps as visible pixel and so appearing in the texture generation.
Even though the FID degradation may seem small, the generated texture maps look significantly worse as shown in Fig. \ref{fig:abl}.
The generated textures include grey pixels (black pixels in the ground-truth appear as grey due to mean shift) due to the misaligned masking and this becomes visually very distracting and unrealistic.
The rendering results of these texture maps are provided in Fig. \ref{fig:mask_examples}.
We find the grey pixels always exist on the generated texture maps, sometimes more subtle as in the second row and
sometimes more significant as in the first row.

We set the best recipe as the baseline and show FIDs of the other methods and their differences with the baseline model in terms of FID metric in the last column.
Baseline model includes our improvements on the data generation but uses the default architectures proposed by \cite{pavllo2020convolutional}.
Next, we test the proposed generative learning pipeline. We see a significant improvement from learnable embeddings fed into the discriminator as shown in Table \ref{tab:abl}.
FID improves from $82.19$ to $66.57$. 
This large improvement is due to the spatially aligned inputs to the discriminator.
Learning the positions of patches is very important when discriminator evaluates their GAN loss.
Discriminator specializes to patches and back propagates better signals to the generator.
The attention mechanism in the generator further improves the FID by  enabling generator to receive signals from distant features.
Instead of position based attention, we also experiment with self-attention but do not observe any improvements.

\subsubsection{Single-view Datasets}
Additionally, we experiment on Pascal 3D+ and CUB datasets.
In our experiments with single-view image datasets, we do not use our improvements on the reconstruction pipeline since ground-truth camera parameters  and silhouttes generated by Mask-RCNN are accurate.
Additionally,  since the model is trained on single view images, the learned meshes overfit to the silhouettes generated by Mask-RCNN.
Therefore, Mask-RCNN mask predictions and 3D projected masks by the renderer are the same.
On this dataset, we only learn textures with ConvMesh and our proposed architecture.
The trend is similar on Pascal 3D+ car and CUB datasets as presented in Table \ref{tab:p3d} and \ref{tab:cub}, respectively.
The discriminator augmented with learned embeddings and generator with position based attention consistently improve the FID results.

\begin{figure*}
    \centering
    \includegraphics[width=1.0\textwidth]{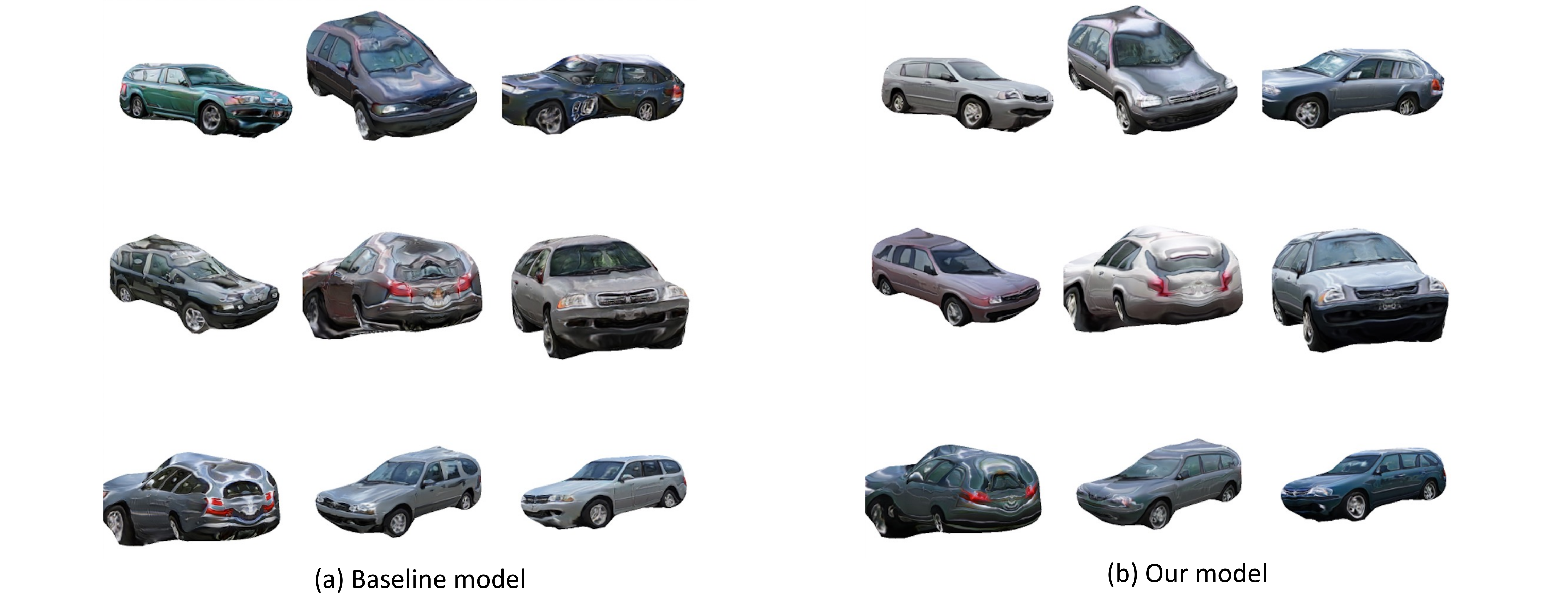}
        \caption{Qualitative results on Pascal 3D+ dataset \cite{xiang2014beyond}. Images are rendered on novel views for baseline and our final models.
        Our model outputs rendered images with less artifacts compared to the baseline model.}
        \label{fig:p3d}
\end{figure*}

\begin{figure*}
    \centering
    \includegraphics[width=1.0\textwidth]{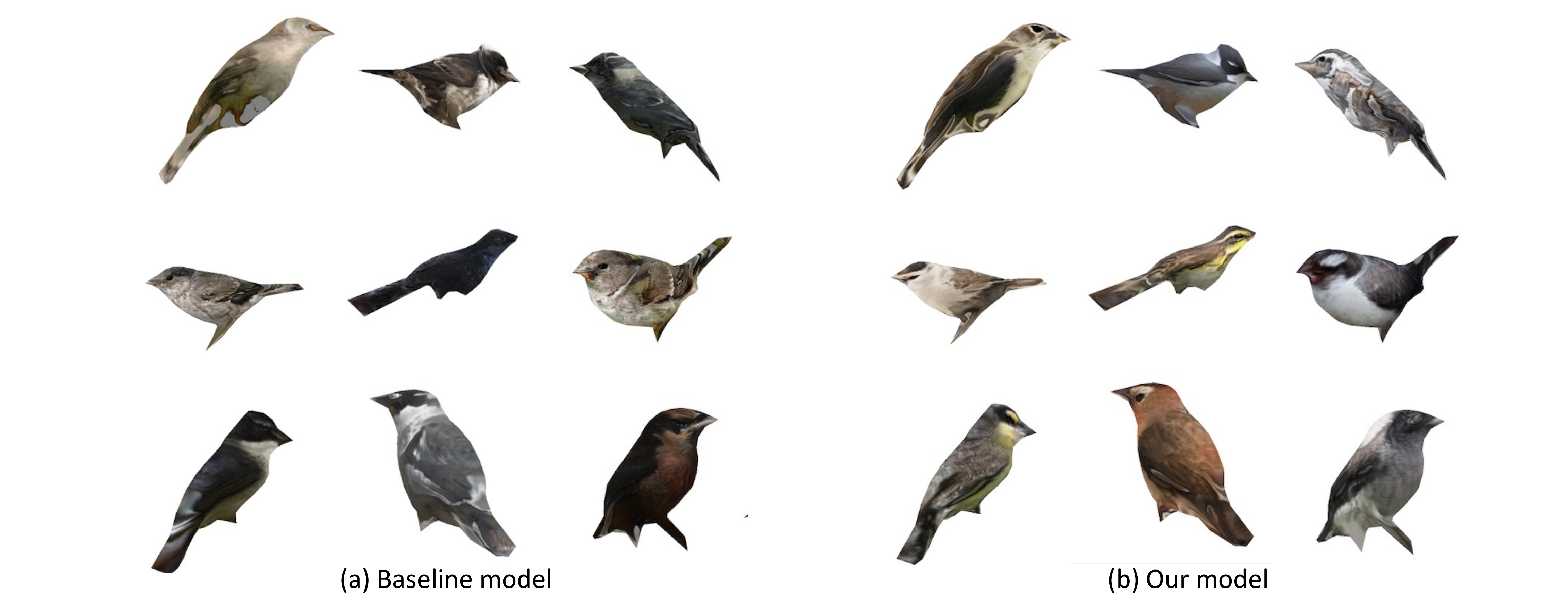}
        \caption{Qualitative results on CUB  dataset \cite{welinder2010caltech}. Images are rendered on novel views for baseline and our final models.
        Our method outputs better results with more diversity and more details compared to the baseline model.}
        \label{fig:cub}
\end{figure*}

\subsection{Qualitative Results}
\label{sec:qual}

\subsubsection{Multi-view Datasets}
Qualitative results of our final pipeline and  other methods are shown in Fig. \ref{fig:results}.
In the first column, we show generated texture maps, and in the rest we show rendering results from different views.
We first visualize results of the  Dib-r model \cite{chen2019learning} which is trained on single view image dataset, Pascal 3D+.
This model and others that are trained on single view images receive no supervision to reconstruct the invisible areas and having difficulties when rendering novel views.
In the second row, we output the inference results of GANVerse3D \cite{zhang2020image} model.
We obtain these results by inferring the models on samples as shown in Fig. \ref{fig:dataset}, and we pick the best results for visualization after inferring on each image from a sequence.
GANVerse3D is trained on a million of multi-view images generated by StyleGAN.
Thanks to the multi-view consistency, the learned shapes are much better than the Dib-r model, however, the results are still missing realistic details and show artifacts on the tires and the muddled coloring.
These two models are not comparable with our work since we adapt our networks to each sequence but we believe they provide important information  to show the difficulties 3D inference models face.
We additionally run experiments with TexRecon \cite{Waechter2014Texturing} and NerfMM \cite{wang2021nerf} algorithms. TexRecon is the closest work that extracts  explicit meshes with texture from multi-view images and builds on structure-from-motion.
NerfMM  does not require known camera parameters and so applicable to our dataset.
However, we find both of those models work very poorly as the sequences are very challenging due to inconsistent lighting across views.

The other results shown in Fig.\ref{fig:results} provide comparisons of models trained on each sequences and therefore comparable with each other.
Third row presents the results of the reconstruction network which is fine-tuned on this particular sequence.
We find the shape estimate to be very accurate since the shape is consistently present in multi-view images.
The same is not true for texture since we present the light and reflections combined under texture, it has variances among different views.
Therefore, when we require the network to output consistent texture predictions, they are not sharp due to the averaging from multiple views (minimizing the averaged reconstruction loss from multiple views).
This results in blurred colors and muddled details, e.g. on the tires.
On the other hand, the GAN framework models the variance in the lighting and reflection of pseudo-ground truth.
It is able to synthesize slightly different textures as shown in Fig. \ref{fig:final_diverse_2}.
This result shows the importance of GAN loss.
Additional GAN loss could also be added when training the reconstruction pipeline, but we find that not improving the results due to the limitations in the rendering pipeline as was also reported in \cite{pavllo2020convolutional}.
We find the 2D generative pipeline proposed in ConvMesh \cite{pavllo2020convolutional} to be very important in rendering realistic details.
Finally, we compare the baseline Convmesh model and our proposed architecture in the last two rows of Fig. \ref{fig:results}.
Our baseline includes our improvements on the data generation pipeline.
Last, our final results which incorporate the changes in the generator and discriminator are presented.
The improvements are visible between the baseline and our final models as was also observed in FID improvements.
Please refer to project page for more examples.

\subsubsection{Single-view Datasets}

Our visual results on Pascal 3D+ and CUB dataset are shown in Fig. \ref{fig:p3d} and \ref{fig:cub}, respectively.
Since in this setting, we do not perform reconstruction but novel generations, this setting is not comparable with Dib-r or GANVerse3D methods.
The setting is also not comparable with Nerf models \cite{wang2021nerf} since dataset contains single-view image collections.
The same camera parameters from the validation sets are loaded for these comparisons.
The generation process is not aware of which camera parameters will be used.
Since models generate novel texture maps, it is not easy to compare the results in a one-to-one setting.
We output 9 textures for baseline and our model without any cherry-picking.
We find that our model outputs less artifacts and more details overall compared to the baseline.
For example zooming into the 3 cars from the first column, our model outputs less artifacts than the baseline model as was also measured in FID.
The top right example from baseline shows a car rendering with three tires.
On the other hand, our model powered with position-based attention in generator and learnable position embedding in discriminator does not output such artifacts. 
Fig. \ref{fig:cub} shows qualitative results on CUB dataset.
Our method again outputs better results with more diversity and more details compared to the baseline model.
Specifically, baseline model generations are missing eyes of the birds for many examples. On the other hand, the textures generated by our model does not have such problem.


\section{Conclusion}
\label{sec:conc}

In this work, we present a method to reconstruct consistent textured 3D models from multi-view images with no annotations by employing pretrained networks.
We pose the reconstruction training as an adaptation problem which starts with learning a category-specific 3D reconstruction model
from a collection of single-view images of the same category that jointly predicts the shape, texture, and camera pose of an image followed by a generative learning pipeline.
Our improvements on the data generation pipeline and network architectures result in significant improvements both qualitatively and quantitatively.
These improvements are not limited to multi-view datasets and also improve results on single-view datasets.
As far as limitations, same as the prior work \cite{kanazawa2018learning, zhang2020image, pavllo2020convolutional}, we deform our final meshes from a sphere and cannot handle objects with holes. Therefore, we cannot model the interior of the car jointly.
Additionally, even though our framework does not require human-annotated labels for novel sequences, it still requires lose estimates on the camera parameters and mask silhouettes which we obtained from pre-trained networks.

\ifCLASSOPTIONcaptionsoff
  \newpage
\fi

{
\bibliographystyle{ieee}
\bibliography{egbib}
}

\end{document}